\documentclass{article}
\usepackage{longtable}
\usepackage{spconf,amsmath,graphicx,booktabs,gensymb}
\usepackage{tabularx,multirow}
\usepackage{pdflscape}
\usepackage{lscape}
\usepackage{rotating}
\usepackage{longtable}
\usepackage{url}
\usepackage{tabularray}

\title{MWIRSTD: A MWIR SMALL TARGET DETECTION DATASET}
\name{Nikhil Kumar$^{\dagger*}$\thanks{*Equal Contribution} \quad  Avinash Upadhyay$^{\ddagger*}$ \quad Shreya Sharma$^{\ddagger\S}$ \quad Manoj Sharma$^{\ddagger}$ \quad
 Pravendra Singh$^{\dagger}$ }
\address{ $^{\dagger}$ IIT Roorkee, $^{\ddagger}$Bennett University, $^{\S}$Visual Cognitional Laboratory Pvt Ltd, \\  $^{\dagger}$ nikhil\_k1@cs.iitr.ac.in
, $^{\ddagger}$ avinres@gmail.com, $^{\ddagger}$ manoj.sharma1@bennett.edu.in,\\ $^{\ddagger}$ e21cseu0601@bennett.edu.in, $^{\dagger}$pravendra.singh@cs.iitr.ac.in}
\begin{document}
%
\maketitle
\begin{abstract}
This paper presents a novel mid-wave infrared (MWIR) small target detection dataset (MWIRSTD) comprising 14 video sequences containing approximately 1053 images with annotated targets of three distinct classes of small objects. Captured using cooled MWIR imagers, the dataset offers a unique opportunity for researchers to develop and evaluate state-of-the-art methods for small object detection in realistic MWIR scenes. Unlike existing datasets, which primarily consist of uncooled thermal images or synthetic data with targets superimposed onto the background or vice versa, MWIRSTD provides authentic MWIR data with diverse targets and environments. Extensive experiments on various traditional methods and deep learning-based techniques for small target detection are performed on the proposed dataset, providing valuable insights into their efficacy. The dataset and code are available at \url{https://github.com/avinres/MWIRSTD.}
\end{abstract}
\begin{keywords}
Small target detection, small and dim target detection, point target detection, MWIR, dataset, defence application 
\end{keywords}
\section{Introduction}
\label{sec:intro}

The electromagnetic spectrum's infrared (IR) band encompasses wavelengths from 700 nm to 1 mm. Table \ref{tab:irspec} outlines the various bands within the IR spectrum commonly used for imaging applications. It's worth noting that while the IR spectrum shares its lower end with the visible spectrum's red edge, research suggests that a considerable portion of the IR spectrum is unsuitable for typical imaging due to poor transmission characteristics \cite{ holst2000common, lloyd2013thermal}.

In both military and civilian realms, the precise detection of small objects in IR imaging holds significant importance. Modern computer vision algorithms leverage deep neural networks and data-driven machine learning techniques to achieve this. These algorithms aim to enhance object detection accuracy while minimizing false alarms. However, the efficacy of such approaches heavily relies on robust datasets for training and validation \cite{hudson1969infrared}. Despite the growing demand for high-quality datasets, publicly available resources often fall short. Many existing datasets are constructed by overlaying pre-recorded IR image backgrounds with superimposed targets. This approach, while helpful, lacks the authenticity and complexity of real-world scenarios.
Addressing this gap, our study utilized a cooled mid-wave infrared (MWIR) imager to capture video sequences featuring small targets in authentic environments. This endeavour resulted in the creation of the first publicly available dataset focused on small targets captured in MWIR. This dataset represents a valuable asset for developing algorithms tailored to defence applications. Furthermore, a noteworthy characteristic of the dataset is that many small targets exhibit spatio-temporal regularity during motion. This inherent feature makes the dataset suitable for object detection and well-suited for tracking purposes.

\begin{table}[t]
\caption{Various bands in IR spectrum for imaging.}
\label{tab:irspec}
\begin{center}
    \begin{tabular}{@{}cc@{}}
\toprule
\textbf{Band}             & \textbf{Wavelength} \\ \midrule
Near Infrared (NIR)       & 0.7um-0.9um         \\
Short Wave Infrared (SWIR) & 0.9um-2.5um         \\
Mid Wave Infrared MWIR)   & 3um-5um             \\
Long Wave Infrared (LWIR)  & 8um-12um            \\ \bottomrule
\end{tabular}
\end{center}
\vspace{-20pt}
\end{table}

 \begin{table*}[!ht]
\caption{Details of various available datasets commonly used in small target detection. Table courtesy~\cite{kumar2023small}.}
\label{tab:dtst}
\resizebox{2\columnwidth}{!}{%
\begin{tabular}{@{}ccccccc@{}}
\toprule
\textbf{Dataset} &
  \textbf{Image Type} &
  \textbf{Background Scene} &
  \textbf{\begin{tabular}[c]{@{}c@{}}Number of\\ Images\end{tabular}} &
  \textbf{Label Type} &
  \textbf{Target Type} &
  \textbf{Availability} \\ \midrule
\textbf{NUAA-SIRST \cite{dai2021asymmetric}} &
  Real &
  Cloud/City/Sea &
  427 &
  \begin{tabular}[c]{@{}c@{}}Manual Coarse\\ Label\end{tabular} &
  Point/Spot/Extended &
  Public \\ \\ 
\textbf{NUST-SIRST \cite{wang2019miss}} &
  Synthetic &
  Cloud/City/River/Road &
  10,000 &
  \begin{tabular}[c]{@{}c@{}}Manual Coarse\\ Label\end{tabular} &
  Point Spot &
  Public \\ \\ 
\textbf{CQU-SIRST \cite{gao2013infrared}} &
  Synthetic &
  Cloud/City/Sea &
  1676 &
  Ground Truth &
  Point Spot &
  Private \\ \\ 
\textbf{NUDT-SIRST \cite{li2022dense}} &
  Synthetic &
  Cloud/City/Sea/Highlight/Field &
  1327 &
  Ground Truth &
  Point/Spot/Extended &
  Public \\ \\ 
\textbf{IRSTD \cite{zhang2022isnet}} &
  Real &
  Sea/River/Field/Mountain/City/Cloud &
  1000 &
  Ground Truth &
  Point/Spot/Extended &
  Public \\ \\ \bottomrule
\end{tabular}
}
\end{table*}

\subsection{Significance of small targets in MWIR imagery}
Due to the high temperatures, military targets like missiles and aircraft plumes exhibit more prominent signatures in the Medium-wave infrared (MWIR) band than in the long-wave infrared (LWIR) band. The MWIR band plays a critical role in IR countermeasure activities. Several electro-optical systems used in aerial countermeasure activities, such as IR Search and Track (IRST) and Missile Approach Warning Systems (MAWS), are designed to operate within this specific range of the electromagnetic spectrum, as discussed in works \cite{de1995irst} and \cite{holm2010missile}. These systems detect potential threats over long distances, sometimes spanning hundreds of kilometres. Targets of military importance are typically large and emit strong infrared radiation. However, they often appear small and faint in images due to their small angular projection on the imaging plane and significant transmission losses. These losses, which involve absorption and scattering \cite{hudson1969infrared}, become increasingly noticeable when travelling long distances. Occasionally, when dealing with long distances, the signatures of certain targets may be limited to just one or two pixels. Even at closer distances, the target can continue to appear small in the image plane.
Unfortunately, it will be a complex task to simultaneously fly multiple aircraft and missiles in order to generate datasets for the purpose of detecting and tracking small targets.
Small targets usually only cover a small portion of the pixels, leading to a scarcity of spatial features. The infrared signatures can vary over time with the changing thermodynamics of the scene and the shifting visual aspect angle of the seen object relative to the position of the imaging device. Infrared imaging frequently experiences significant amounts of noise and background clutter, which may obscure targets, decrease contrast, and lead to a reduced Signal-to-Clutter Ratio (SCR). IR small targets have a lower contrast ratio than small targets in RGB images, resulting in a more considerable similarity to the background. This similarity makes it more difficult to differentiate such targets from their surroundings.
Applying well-known deep learning object detection algorithms such as the RCNN series \cite{girshick2014rich}, YOLO series \cite{redmon2016you}, and SSD \cite{liu2016ssd} for detecting small and dim point targets is not ideal. This is because the pooling layers in these networks may cause the loss of these targets in deeper layers. Researchers have dedicated their efforts to creating specialized deep networks that can effectively identify small infrared targets. 


\vspace{-.2cm}
\section{Related Work}
The proposed work presents a novel Mid-Wave Infrared (MWIR) dataset specifically tailored for small target detection. This section reviews existing datasets relevant to small target detection and provides a comparative analysis of their characteristics.

Several existing datasets play a crucial role in advancing small target detection methodologies. Notable among them are  NUAA-SIRST Dataset \cite{dai2021asymmetric}, NUST-SIRST Dataset \cite{wang2019miss},  CQU-SIRST Dataset \cite{gao2013infrared}, NUDT-SIRST Dataset \cite{li2022dense} and IRSTD Dataset~\cite{zhang2022isnet}. Each of these datasets offers unique challenges and scenarios for algorithm assessment. Table~\ref{tab:dtst} provides a detailed comparison, highlighting key attributes such as dataset size, diversity, and scene complexity across datasets. One of the main limitations of these datasets is that they do not claim to be captured in MWIR. Additionally, many of them have used a target overlay approach on a recorded background, which means they are not representative of real scenarios. On the other hand, this work presents a real dataset captured through a cooled MWIR imager. While these datasets contribute significantly to the field, there remains a need for a specialized real dataset that focuses specifically on the real challenges associated with small target detection within the MWIR spectrum.

In addition to existing datasets, researchers have developed their own distinct datasets to evaluate the performance of their algorithms. Liu et al. \cite{liu2017image} conducted a study involving the creation of a dataset \cite{10249124} aimed at assessing their algorithm's capabilities. Similarly, Kumar et al. \cite{kumar2021detection} developed a dataset tailored to their specific research objectives. These custom datasets enable researchers to tailor scenarios to their algorithm's strengths and weaknesses, providing a more targeted evaluation.

A noteworthy methodology introduced by Naraniya et al. (2021) involves the creation of a specialized dataset through the combination of artificially generated backgrounds with synthetic targets. In this approach, the movement of the target is represented in the NED coordinate system, and Gaussian blurred small targets are superimposed onto backgrounds. This unique dataset creation method adds variability and complexity to the dataset, closely simulating real-world scenarios. While existing small target detection datasets have significantly contributed to advancing target detection algorithms, the proposed work introduces a novel dataset specifically designed for small target detection in MWIR images. 
\begin{table}[h]
\begin{center}
\caption{Details of Imager used for dataset generation.}
\label{tab:imgr}
\begin{tabular}{@{}lll@{}}
\toprule
\textbf{Parameter}      & \textbf{Specification} & \textbf{Remark} \\ \midrule
\textbf{Spectral range} & 3$\mu$m - 5$\mu$m             & MWIR            \\
\textbf{Type}           & Cooled                 & Stirling cycle  \\
\textbf{FPA resolution} & 640 X 512              & InSb            \\
\textbf{FOV (widest)}   & 12 $\degree$ $\times$ 9 $\degree$                & Continuous Zoom \\
\textbf{FOV(narrow)}    & 0.9 $\degree$ $\times$ 0.7 $\degree$              &                 \\ \bottomrule
\end{tabular}
\end{center}
\end{table}
\vspace{-.8cm}
\section{Proposed Dataset}
\subsection{Data Collection}
A MWIR imager, with the specifications outlined in Table \ref{tab:imgr}, was installed on a mountain where the environmental conditions described in Table \ref{tab:env-table} were present. Throughout the exercise, the slant distance between the object and the imager ranged from 300 meters to 950 meters, with the imager mostly in the wide field of view (WFOV).
\vspace{-.5cm}
\subsection{Pre processing}
The imager transmits out the live video in PAL-B format at a frame rate of 25 frames per second. Because of the analog nature of the video output, a frame grabber card was used to convert and record this information in a digital format. Even though it produces a monochrome video sequence, the images are saved as `.jpg' files with three channels due to the frame grabber hardware settings. During the pre-processing stage, the authors extracted sequences with pertinent information from a lengthy video sequence. The dimensions of all extracted frames were resized to 509 $\times$ 655 to exclude areas that do not contain useful information.
 
\vspace{-0.3cm}
\subsection{The Ground truth}
\begin{figure}[h]
	\centering
	\includegraphics[scale=0.7]{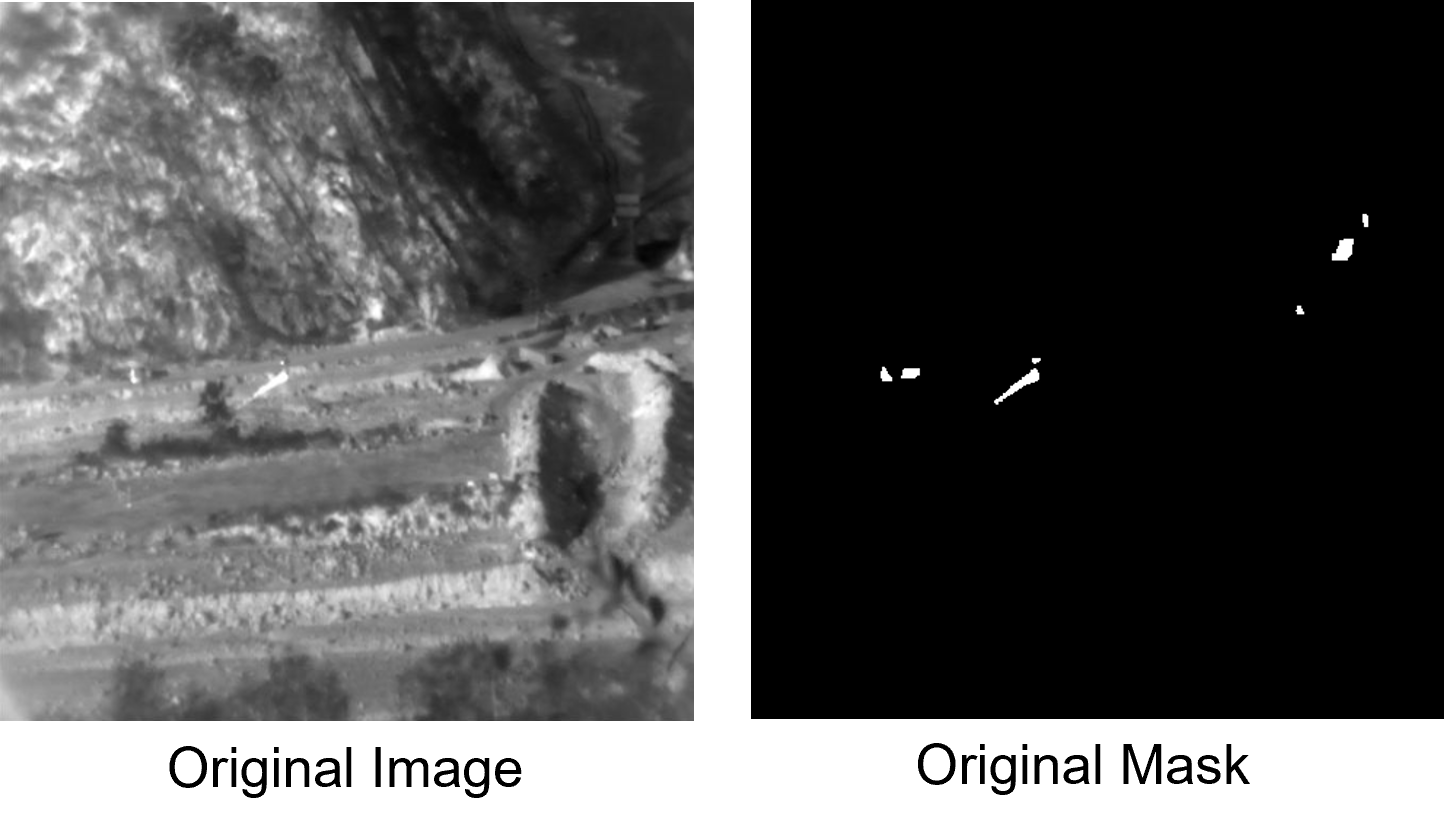}
	\caption{A typical frame of the proposed dataset and its mask.}
	\label{img:frm}
\end{figure}

\begin{figure}[h]
	\centering
	\includegraphics[scale=0.30]{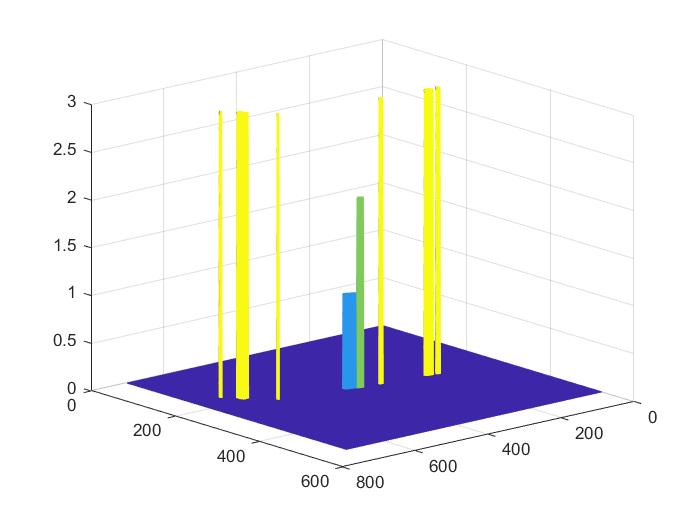}
	\caption{The ground truth of the considered frame shown in Fig.~\ref{img:frm} with annotations for three distinct classes. Class 1 refers to cracker rockets (shown in blue), Class 2 represents debris from cracker rockets (shown in green), and Class 3 includes other small moving targets (shown in yellow). }
	\label{img:gdtr}
\end{figure}

Given that the goal of this setup was to capture small objects in the MWIR range, the authors utilized cracker rockets. In certain situations, a road serves as the background, while multiple moving automobiles function as small targets. The ground truth data includes labels for moving ground vehicles. As shown in Figure \ref{img:frm} and Figure \ref{img:gdtr}, there are three types of labels that are annotated in the ground truth. The first class of annotation has been accomplished for cracker rockets, while the second class of annotation pertains to the debris resulting from these rockets. The third class of annotation is reserved for other moving targets in the scenario, which are often categorized as small targets. Table \ref{tab:decription1} provides a detailed description of the content found in all 14 recorded sequences of the dataset. 

\begin{table}[t]
\begin{center}
\caption{Environmental parameters during dataset collection.}
\label{tab:env-table}
\begin{tabular}{@{}ll@{}}
\toprule
\textbf{Parameter}   & \textbf{Value}          \\ \midrule
\textbf{Wind Speed}  & 6 mph                   \\
\textbf{Humidity}    & 72 $\%$               \\
\textbf{Temperature} & 11.6 $\degree$C - 15 $\degree$ C           \\
\textbf{Time}        & 1400-1700 hrs GMT+0530  \\
\textbf{Visibility}  & Poor (Hazy environment) \\ \bottomrule
\end{tabular}
\end{center}
\end{table}

\begin{table*}[htbp]
  \centering
  \caption{Details of the proposed dataset including sequence number, number of frames, frame size, target details, background composition, and other relevant information.}
    \resizebox{\textwidth}{!}{
    \begin{tabular}{|r|r|p{3.945em}|p{12em}|p{10.335em}|p{11.835em}|}
    \toprule
    \multicolumn{1}{|p{1.72em}|}{\textbf{Seq No }} & \multicolumn{1}{p{3.28em}|}{\textbf{Number of Frames}} & \textbf{Frame Size} & \textbf{Target Details} & \textbf{Background Composition} & \textbf{Additional Information} \\
    \midrule
    1     & 78    & 509x655 (8-bit) & Single Cracker rocket with its debris & Open Field/Road/Mountain/ Vegetation/Houses & Target motion across less, cluttered vegetation and open field \\
    \midrule
    2     & 73    & 509x655 (8-bit) & Single Cracker rocket with its debris & Open Field/Road/Mountain/ Vegetation/Houses & \multicolumn{1}{r|}{} \\
    \midrule
    3     & 63    & 509x655 (8-bit) & Single Cracker rocket with its debris & Open Field/Road/Mountain/ Vegetation/Houses & The target motion also encompasses a densely cluttered region. Almost vertical path. Almost complete FOV is covered by the target. \\
    \midrule
    4     & 65    & 509x655 (8-bit) & Single Cracker rocket with its debris & Open Field/Road/Mountain/ Vegetation/Houses & The target trajectory covers both open field as well as densely cluttered region. Almost vertical path. \\
    \midrule
    5     & 119   & 509x655 (8-bit) & Single Cracker rocket with its debris & Open Field/Road/Mountain/ Vegetation/Houses & Target motion across less cluttered vegetation and the open field \\
    \midrule
    6     & 93    & 509x655 (8-bit) & Single Cracker rocket with its debris & Open Field/Road/Mountain/ Vegetation/Houses & \multicolumn{1}{r|}{} \\
    \midrule
    7     & 74    & 509x655 (8-bit) & Single Cracker rocket with its debris, Multiple moving Vehicles as small targets & Open Field/Road/Mountain/ Vegetation/Houses & Target motion across deep cluttered vegetation. \\
    \midrule
    8     & 56    & 509x655 (8-bit) & Single Cracker rocket with its debris, Multiple moving Vehicles as small targets & Open Field/Road/Mountain/ Vegetation/Houses & \multicolumn{1}{r|}{} \\
    \midrule
    9     & 83    & 509x655 (8-bit) & Single Cracker rocket with its debris, Multiple moving Vehicles as small targets & Open Field/Road/Mountain/ Vegetation/Houses & \multicolumn{1}{r|}{} \\
    \midrule
    10    & 86    & 509x655 (8-bit) & Single Cracker rocket with its debris, Multiple moving Vehicles as small targets & Open Field/Road/Mountain/ Vegetation/Houses & Parabolic trajectory  \\
    \midrule
    11    & 77    & 509x655 (8-bit) & Single Cracker rocket with its debris, Multiple moving Vehicles as small targets & Open Field/Road/Mountain/ Vegetation/Houses & Almost Horizontal Trajectory  \\
    \midrule
    12    & 53    & 509x655 (8-bit) & Single Cracker rocket with its debris, Multiple moving Vehicles, pedestrians, two-wheeler inside FOV & Dense Forest / Road  & Almost Horizontal Trajectory  \\
   
    \bottomrule
    \end{tabular}%
}
  \label{tab:decription1}%
\end{table*}%

\begin{table*}[]
  \centering
 
    \resizebox{\textwidth}{!}{
    \begin{tabular}{|r|r|p{3.945em}|p{12em}|p{10.335em}|p{11.835em}|}
    \toprule
    \multicolumn{1}{|p{1.72em}|}{\textbf{Seq No }} & \multicolumn{1}{p{3.28em}|}{\textbf{Number of Frames}} & \textbf{Frame Size} & \textbf{Target Details} & \textbf{Background Composition} & \textbf{Additional Information} \\
    \midrule
    13    & 75    & 509x655 (8-bit) & Single Cracker rocket with its debris, Multiple moving Vehicles, pedestrians, two-wheeler inside FOV & Dense Forest / Road  & \multicolumn{1}{r|}{} \\
    \midrule
    14    & 68    & 509x655 (8-bit) & Single Cracker rocket with its debris, Multiple moving Vehicles, pedestrians, two-wheeler inside FOV & Dense Forest / Road  & Trajectory at 45 degrees  \\
    \bottomrule
    \end{tabular}%
}
  \label{tab:decription2}%
\end{table*}%




\vspace{-.3cm}
\section{Experiments}
\subsection{Evaluation Metric}

The proposed dataset was evaluated using two widely used metrics, i.e., IoU (Intersection over Union) and PD (Probability of Detection) at the pixel level, to assess the effectiveness of different image processing based computation and deep learning methods. Further, the FAR (False alarm Rate) at the pixel level was also used to provide a more detailed analysis of the results. We have selected these metrics as they are commonly used in target segmentation scenarios and provide a balanced measure of both detection accuracy and precision. Specifically, IoU evaluates the overlap between the predicted bounding box and the ground truth, while PD assesses the ratio of correctly detected pixels as the target to the total number of pixels detected as a target. Furthermore, calculating FAR helps identify any potential issues with false positives, which can be critical in applications where accurate detection is crucial. 

\begin{figure}
    \centering
    \includegraphics[width=0.9\linewidth]{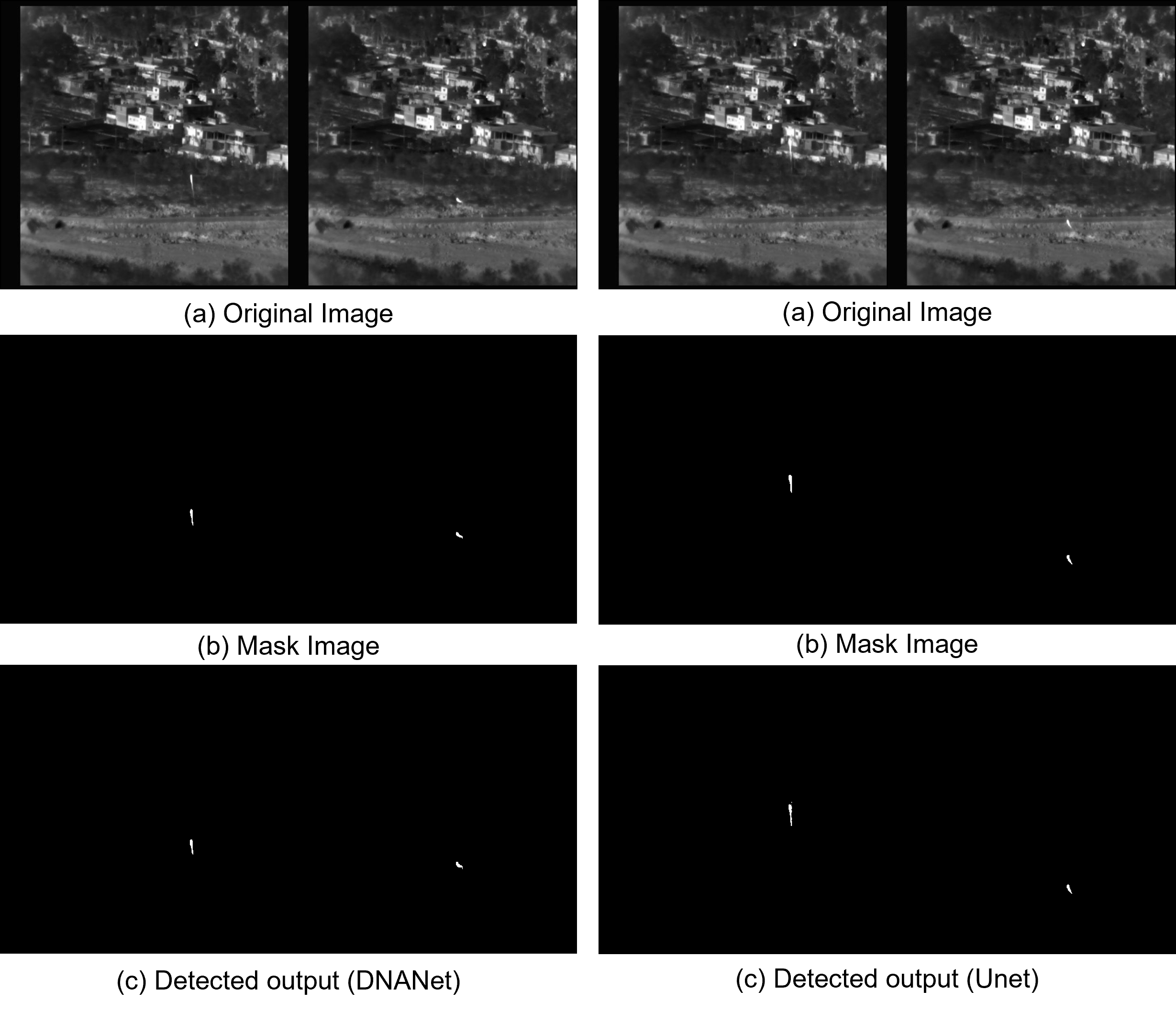}
    \caption{Visual results obtained using Deep learning-based methods.}
    \label{fig:results_deep_learning}
\end{figure}

\begin{figure*}
    \centering
    \includegraphics[width=0.9\linewidth]{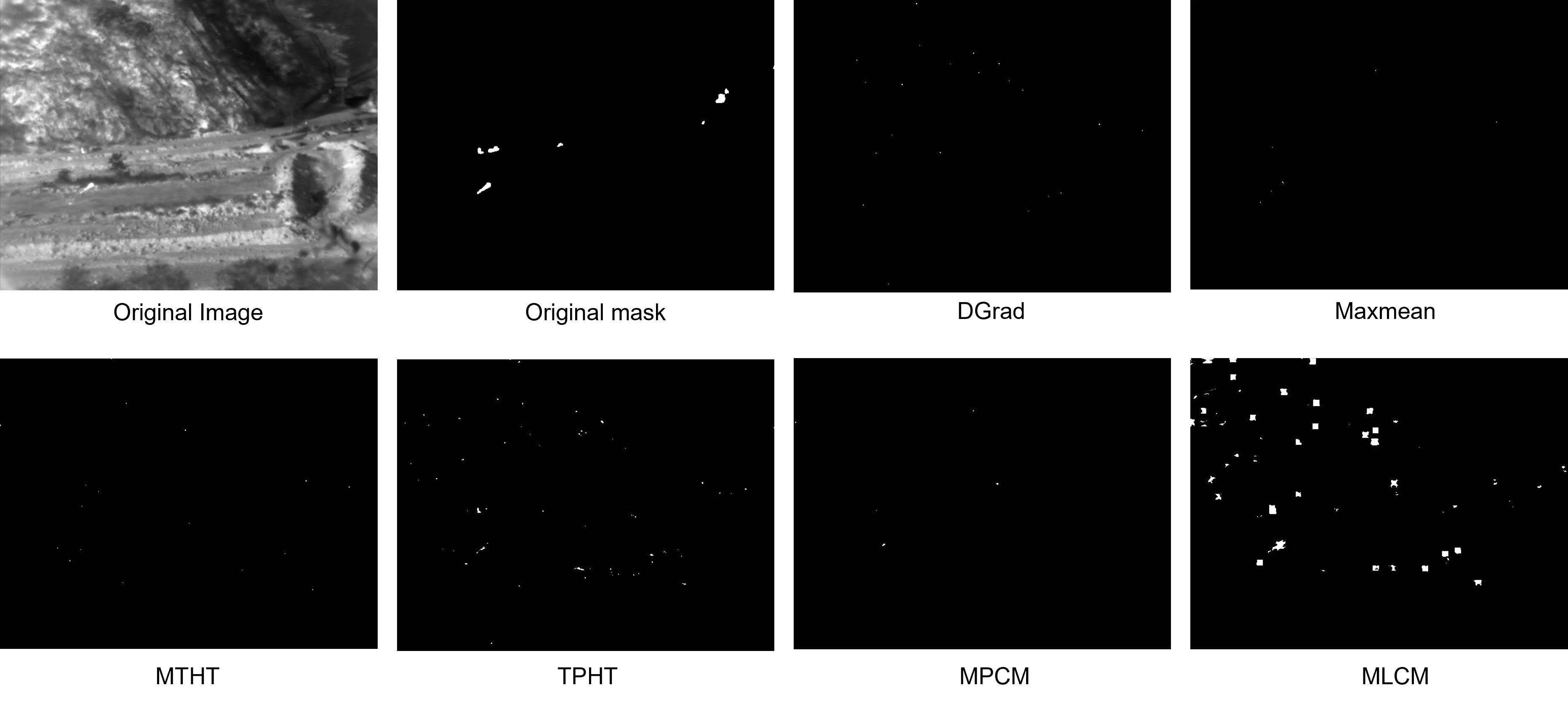}
    \caption{Visual results obtained using traditional computational methods.}
    \label{fig:enter-label}
\end{figure*}

\begin{figure}
    \centering
    \includegraphics[width=1\linewidth]{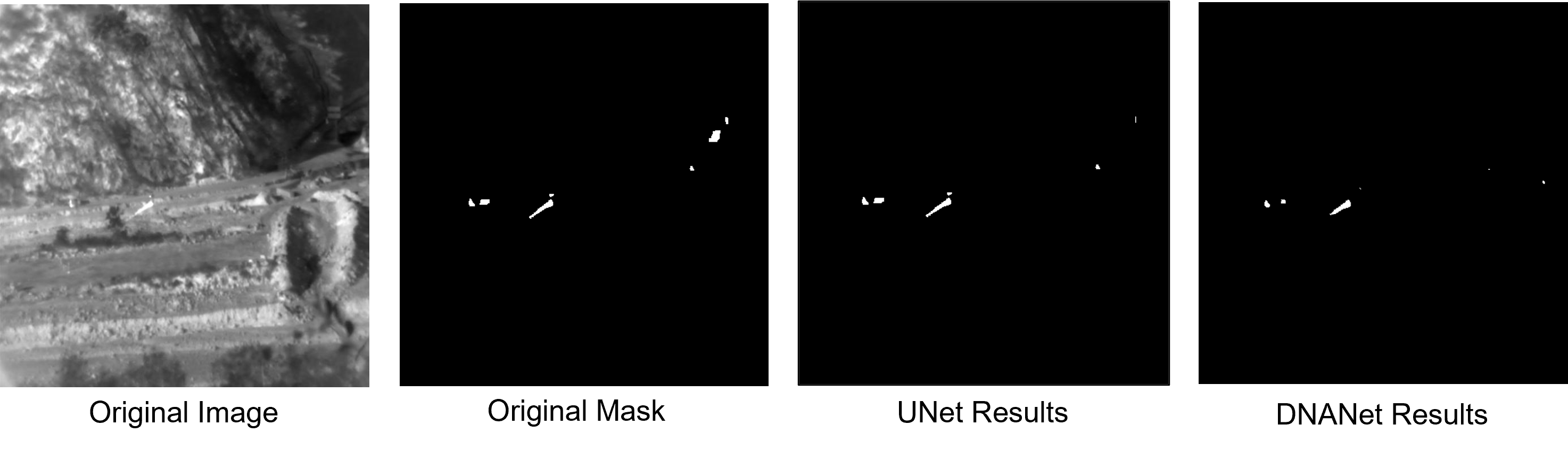}
   \caption{Failed cases for Deep learning-based models.}
    \label{fig:failed}
\end{figure}

\subsection{Results and Analysis}
In this section, we present the evaluation results of six traditional small object detection methods: Top-hat Morphology (TPHT) \cite{gonzalez2004digital}, Modified Top Hat Morphology (MTHM) \cite{bai2008new, bai2010enhancement}, Maxmean \cite{gu2010kernel}, MLCM \cite{6479296}, MRCM \cite{wei2016multiscale}, and DGRAD \cite{wu2020double}) and two state-of-the-art deep learning-based methods U-Net \cite{ronneberger2015u}, DNA-Net \cite{li2022dense} on our proposed MWIR dataset. We employed two commonly used metrics, i.e., mean Intersection over Union (IoU) and mean Probability of Detection (PD), to evaluate the performance of these methods. Additionally, we calculated the mean False Alarm Rate (FAR) at the pixel level to analyze the results further. The deep-learning-based models were trained from scratch on our datasets. Initially, IoU, Pd, and FAR are calculated on a frame basis, and then their mean value for the complete test set has been presented here. 
TPHT is based on conventional top hat morphological operation. In this case, a 5$\times$5 square structuring element has been taken. MTHM is a modified version of classical top hat morphology, and in this, a ring-type structuring element has been considered in place of the square structuring element in the previous case. Maxmean is a statistical algorithm, and while implementing this, the Maxmean operation is implemented on 5$\times$5 window of the image. MLCM, MRCM, and DGRAD use image contrast as a primary feature and define some handcrafted feature extraction operations; based on these operations, any particular pixel is declared as part of the foreground or background. One of the standard requirements for all of the above approaches is the segmentation of the detection plane, and for this activity, the following threshold has been chosen as \[Threshold= Image mean + 0.5*(Image max - Image mean)\]

As observed from Table~\ref{tab:perf}, all the traditional methods exhibit suboptimal performance on our dataset. In contrast, the deep learning-based methods demonstrated exceptional performance, with the U-Net method attaining an average IoU of 0.79 and PD of 0.1124. These findings indicate that deep learning techniques surpass traditional computational approaches in detecting small objects in MWIR images.

Table~\ref{tab:perf} demonstrates the ability of the deep learning-based methods to accurately detect and segment small objects in MWIR images, while the traditional computational methods often produce fragmented or incomplete detections. Overall, our results suggest that deep learning-based methods are superior to traditional computational approaches for small object detection in MWIR imagery. Fig.~\ref{fig:results_deep_learning}, Fig.~\ref{fig:enter-label}, and Fig.~\ref{fig:failed} present a glimpse of visual results obtained using the proposed dataset.
 
\begin{table}[]
\begin{center}

\caption{Performance benchmark evaluation using several traditional and deep learning-based approaches over the proposed dataset.}
\label{tab:perf}
\begin{tabular}{@{}llll@{}}
\toprule
\textbf{Algorithm} & \textbf{IoU} & \textbf{POD} & \textbf{FAR} \\
\textbf{Maxmean}   & 0.0036       & 0.018        & 0.9931       \\
\textbf{THM}       & 0.002        & 0.024        & 0.997        \\
\textbf{MTHM}      & 0.0037       & 0.0164       & 0.992        \\
\textbf{MLCM}      & 0.0027       & 0.046        & 0.996        \\
\textbf{MPCM}      & 0.0033       & 0.007        & 0.9858       \\
\textbf{DGRAD}     & 0.0031       & 0.007        & 0.9829       \\
\textbf{U-NET}     & 0.7945       & 0.8855        & 0.1124       \\
\textbf{DNA-NET}   & 0.6440       & 0.7330        & 0.1569  \\ \bottomrule

\end{tabular}
\end{center}
\end{table}

\section{CONCLUSION}
This dataset, the first ever to capture MWIR with real small targets and real backgrounds, will provide a strong foundation for researchers in this field. Currently, the dataset comprises 14 sequences with a diverse range of foreground and background combinations, making it an ideal testbed for training and validating advanced deep-learning models for detecting small and dim targets. A benchmarking study has revealed that techniques employing deep learning models demonstrate exceptional performance, exhibiting a higher probability of detection while maintaining a lower false alarm rate. The targets in these sequences exhibit a consistent spatiotemporal regularity pattern, suggesting potential for future expansion of this dataset to address tracking-related challenges.

\bibliographystyle{IEEEbib}
\bibliography{icip_det}

\end{document}